\documentclass[12pt]{article}
\usepackage{times}

\usepackage[dvips]{graphicx}
\usepackage{epsfig}
\usepackage{latexsym}
\usepackage{xspace}

\usepackage{amsmath}
\usepackage{amsthm}
\usepackage{amssymb}
\usepackage{color}
\usepackage{algorithm}
\usepackage{algorithmic}

\renewcommand{\(}{\left(}
\renewcommand{\)}{\right)}

\renewcommand{\[}{\left[}
\renewcommand{\]}{\right]}

\newcommand{\U}{{\cal U}}

\newcommand{\Y}{{\cal Y}}

\newcommand{\calS}{{\cal S}}

\DeclareMathOperator{\OLPOMDP}{OLPOMDP}

\newcommand{\olpomdp}{{\mbox{\small$\OLPOMDP$}}}

\newcommand{\Expect}{\mathbold{E}}

\newcommand{\Ex}{\Expect}

\newcommand{\mathbold}[1]{\mbox{\boldmath $\bf#1$}}

\renewcommand{\Re}{{\mathbb R}}

\usepackage{color}

\definecolor{darkred}{rgb}{0.7,0.2,0.2}

\definecolor{bgblue}{rgb}{0.04,0.39,0.53}

\newtheorem{theorem}{Theorem}

\allowdisplaybreaks[1]

\title{Hebbian Synaptic Modifications in Spiking Neurons that Learn}

\author{Peter L.~Bartlett and Jonathan Baxter\\
Research School of Information Sciences and Engineering \\
Australian National University\\
Peter.Bartlett@anu.edu.au, Jonathan.Baxter@anu.edu.au}
\date{November 27, 1999}
\begin{document}

\maketitle

\begin{abstract} 

 In this paper, we derive a new model of synaptic plasticity, based on
 recent algorithms for reinforcement learning (in which an agent
 attempts to learn appropriate actions to maximize its long-term
 average reward).  We show that these direct reinforcement learning
 algorithms also give locally optimal performance for the problem of
 reinforcement learning with multiple agents, without any explicit
 communication between agents.  By considering a network of spiking
 neurons as a collection of agents attempting to maximize the
 long-term average of a reward signal, we derive a synaptic update
 rule that is qualitatively similar to Hebb's postulate.  This rule
 requires only simple computations, such as addition and leaky
 integration, and involves only quantities that are available in the
 vicinity of the synapse.  Furthermore, it leads to synaptic
 connection strengths that give locally optimal values of the long
 term average reward.  The reinforcement learning paradigm is
 sufficiently broad to encompass many learning problems that are
 solved by the brain. We illustrate, with simulations, that the
 approach is effective for simple pattern classification and motor
 learning tasks.

\end{abstract}

\section{What is a good synaptic update rule?}

It is widely accepted that the functions performed by neural circuits
are modified by adjustments to the strength of the synaptic connections
between neurons. In the 1940s, Donald Hebb speculated that such
adjustments are associated with simultaneous (or nearly simultaneous)
firing of the presynaptic and postsynaptic neurons~\cite{hebb}:
	\begin{quotation}
	 When an axon of cell $A$ ... persistently takes part in firing
	 [cell $B$], some growth process or metabolic change takes place
	 [to increase] $A$'s efficacy as one of the cells firing $B$.
	\end{quotation}
Although this postulate is rather vague, it provides the
important suggestion that the computations performed by neural
circuits could be modified by a simple cellular mechanism.
Many candidates for Hebbian synaptic update rules have been
suggested, and there is considerable experimental evidence of
such mechanisms (see, for instance,~\cite{BlissLomo73,%
StantonSejnowski89,LoPoo91,MageeJohnston97,Markram97,BiPoo98}).

Hebbian modifications to synaptic strengths seem intuitively reasonable
as a mechanism for modifying the function of a neural circuit. However, it
is not clear that these synaptic updates actually improve the performance
of a neural circuit in any useful sense.  Indeed, simulation studies of
specific Hebbian update rules have illustrated some serious shortcomings
(see, for example,~\cite{medina99}).

In contrast with the ``plausibility of cellular mechanisms'' approach,
most artificial neural network research has emphasized performance in
practical applications.  Synaptic update rules for artificial neural
networks have been devised that minimize a suitable cost function.
Update rules such as the backpropagation algorithm~\cite{Retal}
(see~\cite{fine99} for a more detailed treatment) perform gradient
descent in parameter space: they modify the connection strengths in a
direction that maximally decreases the cost function, and hence leads
to a local minimum of that function. Through appropriate choice of the
cost function, these parameter optimization algorithms have allowed
artificial neural networks to be applied (with considerable success)
to a variety of pattern recognition and predictive modelling problems.

Unfortunately, there is little evidence that the (rather complicated)
computations required for the synaptic update rule in parameter
optimization procedures like the backpropagation algorithm can be
performed in biological neural circuits.  In particular, these algorithms
require gradient signals to be propagated backwards through the network.

This paper presents a synaptic update rule that provably optimizes the
performance of a neural network, but requires only simple computations
involving signals that are readily available in biological neurons. This
synaptic update rule is consistent with Hebb's postulate.

Related update rules have been proposed in the past.  For instance,
the updates used in the adaptive search elements (ASEs) described
in~\cite{BarSutBrou81,BarSut81,BarAndSut81,BarSutAnd83} are of a similar
form (see also~\cite{tesauro86}).  However, it is not known in what sense
these update rules optimize performance.  The update rule we present here
is based on similar foundations to the REINFORCE class of algorithms
introduced by Williams~\cite{williams92}.  However, when applied to
spiking neurons such as those described here, REINFORCE leads to parameter
updates in the steepest ascent direction in two limited situations: when
the reward depends only on the current input to the neuron and the neuron
outputs do not affect the statistical properties of the inputs, and when
the reward depends only on the sequence of inputs since the arrival of the
last reward value. Furthermore, in both cases the parameter updates must
be carefully synchronized with the timing of the reward values, which is
especially problematic for networks with more than one layer of neurons.

In Section~\ref{section:drl}, we describe {\it reinforcement learning
problems\/}, in which an agent aims to maximize the long-term
average of a {\it reward signal\/}.  Reinforcement learning is a
useful abstraction that encompasses many diverse learning problems,
such as supervised learning for pattern classification or predictive
modelling, time series prediction, adaptive control, and game playing.
We review the {\it direct reinforcement learning algorithm\/} we proposed
in~\cite{jair_01a} and show in Section~\ref{section:multi-drl} that, in the
case of multiple independent agents cooperating to optimize performance,
the algorithm conveniently decomposes in such a way that the agents are
able to learn independently with no need for explicit communication.

In Section~\ref{section:drl-nn}, we consider a network of model neurons
as a collection of agents cooperating to solve a reinforcement learning
problem, and show that the direct reinforcement learning algorithm
leads to a simple synaptic update rule, and that the decomposition
property implies that only local information is needed for the updates.
Section~\ref{section:mechanisms} discusses possible mechanisms for the
synaptic update rule in biological neural networks.

The parsimony of requiring only one simple mechanism to optimize
parameters for many diverse learning problems is appealing
(cf~\cite{valiant94}).  In Section~\ref{section:simulation}, we present
results of simulation experiments, illustrating the performance of this
update rule for pattern recognition and adaptive control problems.

\section{Reinforcement learning}
\label{section:drl}

`Reinforcement learning' refers to a general class of learning problems
in which an agent attempts to improve its performance at some task. For
instance, we might want a robot to sweep the floor of an office; to
guide the robot, we provide feedback in the form of occasional rewards,
perhaps depending on how much dust remains on the floor.  This section explains
how we can formally define this class of problems and shows that it
includes as special cases many conventional learning problems. It also
reviews a general-purpose learning method for reinforcement learning
problems.

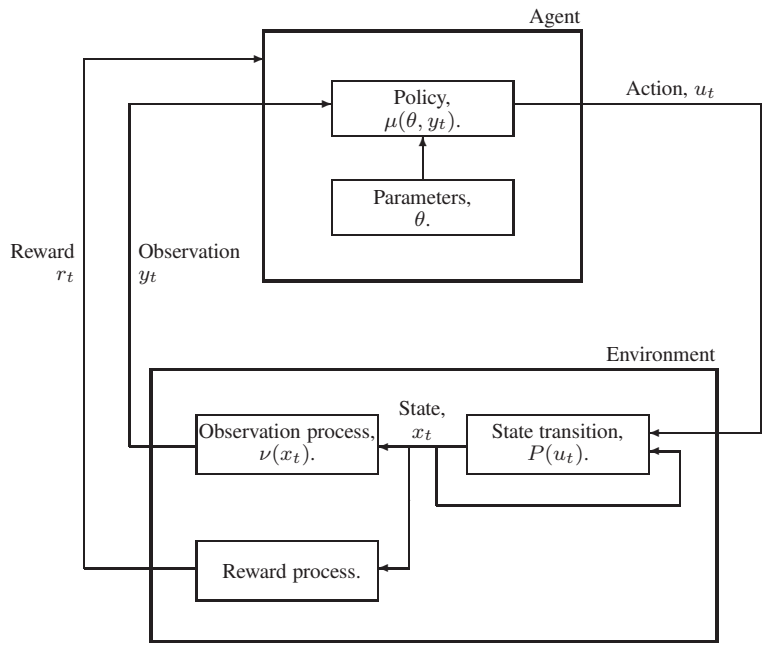
\begin{figure}
\center{\setlength{\unitlength}{2486sp}%
\begingroup\makeatletter\ifx\SetFigFont\undefined%
\gdef\SetFigFont#1#2#3#4#5{%
  \reset@font\fontsize{#1}{#2pt}%
  \fontfamily{#3}\fontseries{#4}\fontshape{#5}%
  \selectfont}%
\fi\endgroup%
\begin{picture}(7452,6342)(211,-6844)
\put(1441,-2986){\makebox(0,0)[lb]{\smash{\SetFigFont{8}{9.6}{\familydefault}{\mddefault}{\updefault}Observation}}}
\put(1441,-3211){\makebox(0,0)[lb]{\smash{\SetFigFont{8}{9.6}{\familydefault}{\mddefault}{\updefault}$y_t$}}}
\put(811,-2986){\makebox(0,0)[rb]{\smash{\SetFigFont{8}{9.6}{\familydefault}{\mddefault}{\updefault}Reward}}}
\put(811,-3211){\makebox(0,0)[rb]{\smash{\SetFigFont{8}{9.6}{\familydefault}{\mddefault}{\updefault}$r_t$}}}
\put(5626,-4786){\makebox(0,0)[b]{\smash{\SetFigFont{8}{9.6}{\familydefault}{\mddefault}{\updefault}State transition,}}}
\put(5626,-5011){\makebox(0,0)[b]{\smash{\SetFigFont{8}{9.6}{\familydefault}{\mddefault}{\updefault}$P(u_t)$.}}}
\thinlines
\put(4726,-5146){\framebox(1800,585){}}
\put(4276,-4561){\makebox(0,0)[b]{\smash{\SetFigFont{8}{9.6}{\familydefault}{\mddefault}{\updefault}State,}}}
\put(4276,-4786){\makebox(0,0)[b]{\smash{\SetFigFont{8}{9.6}{\familydefault}{\mddefault}{\updefault}$x_t$}}}
\put(2926,-4786){\makebox(0,0)[b]{\smash{\SetFigFont{8}{9.6}{\familydefault}{\mddefault}{\updefault}Observation process,}}}
\put(2926,-5011){\makebox(0,0)[b]{\smash{\SetFigFont{8}{9.6}{\familydefault}{\mddefault}{\updefault}$\nu(x_t)$.}}}
\put(2026,-5146){\framebox(1800,585){}}
\put(2026,-6406){\framebox(1800,585){}}
\put(2971,-6181){\makebox(0,0)[b]{\smash{\SetFigFont{8}{9.6}{\familydefault}{\mddefault}{\updefault}Reward process.}}}
\put(4726,-4876){\vector(-1, 0){900}}
\put(4141,-4876){\line( 0,-1){1215}}
\put(4141,-6091){\vector(-1, 0){315}}
\put(4411,-4876){\line( 0,-1){585}}
\put(4411,-5461){\line( 1, 0){2430}}
\put(6841,-5461){\line( 0, 1){540}}
\put(6841,-4921){\vector(-1, 0){315}}
\thicklines
\put(1576,-6811){\framebox(5625,2700){}}
\put(7201,-4021){\makebox(0,0)[rb]{\smash{\SetFigFont{8}{9.6}{\familydefault}{\mddefault}{\updefault}Environment}}}
\put(4276,-2446){\makebox(0,0)[b]{\smash{\SetFigFont{8}{9.6}{\familydefault}{\mddefault}{\updefault}Parameters,}}}
\put(4276,-2671){\makebox(0,0)[b]{\smash{\SetFigFont{8}{9.6}{\familydefault}{\mddefault}{\updefault}$\theta$.}}}
\thinlines
\put(3376,-2761){\framebox(1800,540){}}
\put(4276,-1456){\makebox(0,0)[b]{\smash{\SetFigFont{8}{9.6}{\familydefault}{\mddefault}{\updefault}Policy,}}}
\put(4276,-1681){\makebox(0,0)[b]{\smash{\SetFigFont{8}{9.6}{\familydefault}{\mddefault}{\updefault}$\mu(\theta,y_t)$.}}}
\put(3376,-1771){\framebox(1800,540){}}
\put(4276,-2221){\vector( 0, 1){450}}
\thicklines
\put(2701,-3211){\framebox(3150,2475){}}
\put(5851,-646){\makebox(0,0)[rb]{\smash{\SetFigFont{8}{9.6}{\familydefault}{\mddefault}{\updefault}Agent}}}
\thinlines
\put(4141,-5416){\makebox(2.6458,18.5208){\SetFigFont{5}{6}{\rmdefault}{\mddefault}{\updefault}.}}
\put(2026,-4876){\line(-1, 0){675}}
\put(1351,-4876){\line( 0, 1){3420}}
\put(1351,-1456){\vector( 1, 0){2025}}
\put(2026,-6091){\line(-1, 0){1125}}
\put(901,-6091){\line( 0, 1){5085}}
\put(901,-1006){\vector( 1, 0){1800}}
\put(5176,-1456){\line( 1, 0){2475}}
\put(7651,-1456){\line( 0,-1){3285}}
\put(7651,-4741){\vector(-1, 0){1125}}
\put(6301,-1366){\makebox(0,0)[lb]{\smash{\SetFigFont{8}{9.6}{\familydefault}{\mddefault}{\updefault}Action, $u_t$}}}
\end{picture}}
\caption{\label{figure:pomdp}%
Partially observable Markov decision process (POMDP).}
\end{figure}

We can model the interactions between an agent and its environment
mathematically as a {\it partially observable Markov decision process}
(POMDP).  Figure~\ref{figure:pomdp} illustrates the features of a
POMDP.  At each (discrete) time step $t$, the agent and the
environment are in a particular {\it state} $x_t$ in a state space
$\calS$. For our cleaning robot, $x_t$ might include the agent's
location and orientation, together with the location of dust and
obstacles in the office.  The state at time $t$ determines an {\it
observation vector} $y_t$ (from some set $\Y$) that is seen by the
agent. For instance in the cleaning example, $y_t$ might consist of
visual information available at the agent's current location. Since
observations are typically noisy, the relationship between the state
and the corresponding observation is modelled as a probability
distribution $\nu(x_t)$ over observation vectors. Notice that the
probability distribution depends on the state.

When the agent sees an observation vector $y_t$, it decides on an {\it
action} $u_t$ from some set $\U$ of available actions. In the office
cleaning example, the available actions might consist of directions in
which to move or operations of the robot's broom.

A mapping from observations to actions is referred to as a {\it policy}.
We allow the agent to choose actions using a {\it randomized policy}.
That is, the observation vector $y_t$ determines a probability
distribution $\mu(y_t)$ over actions, and the action is chosen randomly
according to this distribution. We are concerned with randomized policies
that depend on a vector $\theta\in\Re^k$ of $k$ {\it parameters} (and
we write the probability distribution over actions as $\mu(y_t,\theta)$).

The agent's actions determine the evolution of states, possibly in a
probabilistic way. To model this, each action determines the
probabilities of transitions from the current state to possible
subsequent states. For a finite state space $\calS$, we can write
these probabilities as a transition probability matrix,
$P(u_t)$. Here, the $i,j$-the entry of $P(u_t)$ ($p_{ij}(u_t)$) is the
probability of making a transition from state $i$ to state $j$ given
that the agent took action $u_t$ in state $i$.  In the office, the actions
chosen by the agent determine its location and orientation and the
location of dust and obstacles at the next time instant, perhaps with
some random element to model the probability that the agent slips or
bumps into an obstacle.

Finally, in every state, the agent receives a reward signal $r_t$, which
is a real number. For the cleaning agent, the reward might be zero most
of the time, but take a positive value when the agent removes some dust.

The aim of the agent is to choose a policy (that is, the parameters
that determine the policy) so as to maximize the long-term average of
the reward,
 \begin{equation}\label{equation:eta}
  \eta = \lim_{T\to\infty} \Ex \[ \frac{1}{T} \sum_{t=1}^T r_t \].
 \end{equation}
(Here, $\Ex$ is the expectation operator.)  This problem is made more
difficult by the limited information that is available to the agent.
We assume that at each time step the agent sees only the observations
$y_t$ and the reward $r_t$ (and is aware of its policy and the actions
$u_t$ that it chooses to take).  It has no knowledge of the underlying
state space, how the actions affect the evolution of states, how the
reward signals depend on the states, or how the observations depend on
the states.

\subsection{Other learning tasks viewed as reinforcement learning}

Clearly, the reinforcement learning problem described above provides
a good model of adaptive control problems, such as the acquisition of
motor skills. However, the class of reinforcement learning problems is
broad, and includes a number of other learning problems that are solved
by the brain.  For instance, the {\em supervised learning} problems of
pattern recognition and predictive modelling require labels (such as
an appropriate classification) to be associated with patterns. These
problems can be viewed as reinforcement learning problems with reward
signals that depend on the accuracy of each predicted label.  {\em Time
series prediction,} the problem of predicting the next item in a sequence,
can be viewed in the same way, with a reward signal that corresponds to
the accuracy of the prediction.  More general {\em filtering problems}
can also be viewed in this way. It follows that a single mechanism for
reinforcement learning would suffice for the solution of a considerable
variety of learning problems.

\subsection{Direct reinforcement learning}

A general approach to reinforcement learning problems was presented
recently in~\cite{jair_01a,jair_01b}. Those papers considered agents that use
parameterized policies, and introduced general-purpose reinforcement
learning algorithms that adjust the parameters in the direction that
maximally increases the average reward. Such algorithms converge to
policies that are locally optimal, in the sense that any further
adjustment to the parameters in any direction cannot improve the
policy's performance. This section reviews the algorithms introduced
in~\cite{jair_01a,jair_01b}. The next two sections show how these
algorithms can be applied to networks of spiking neurons.

The {\it direct reinforcement learning} approach presented
in~\cite{jair_01a}, building on ideas due to a number of
authors~\cite{williams92,cao97,cao98,kimura97,marbach98}, adjusts the
parameters $\theta$ of a randomized policy that, on being presented with
the observation vector $y_t$, chooses actions according to a probability
distribution $\mu(y_t,\theta)$.  The approach involves the computation
of a vector $z_t$ of $k$ real numbers (one component for each component
of the parameter vector $\theta$) that is updated according to
  \begin{equation}\label{equation:zupdate}
    z_{t+1} = \beta z_t +
      \frac{\nabla \mu_{u_t}(y_t,\theta)}{\mu_{u_t}(y_t,\theta)},
  \end{equation}
where $\beta$ is a real number between $0$ and $1$,
$\mu_{u_t}(y_t,\theta)$ is the probability of the action $u_t$ under
the current policy, and $\nabla$ denotes the gradient with respect to
the parameters $\theta$ (so $\nabla \mu_{u_t}(y_t,\theta)$ is a vector
of $k$ partial derivatives).  The vector $z_t$ is used to update the
parameters, and can be thought of as an average of the `good' directions
in parameter space in which to adjust the parameters if a large value
of reward occurs at time $t$.  The first term in the right-hand-side of
\eqref{equation:zupdate} ensures that $z_t$ remembers past values of the
second term.  The numerator in the second term is in the direction in
parameter space which leads to the maximal increase of the probability
of the action $u_t$ taken at time $t$. This direction is divided by the
probability of $u_t$ to ensure more ``popular'' actions don't end up
dominating the overall update direction for the parameters. Updates to
the parameters correspond to weighted sums of these normalized directions,
where the weighting depends on future values of the reward signal.

Theorems~3 and~6 in~\cite{jair_01a} show that if $\theta$ remains constant,
the long-term average of the product $r_t z_t$ is a good approximation
to the gradient of the average reward with respect to the parameters,
provided $\beta$ is sufficiently close to $1$. It is clear from
Equation~(\ref{equation:zupdate}) that as $\beta$ gets closer to
$1$, $z_t$ depends on measurements further back in time.  Theorem~4
in~\cite{jair_01a} shows that, for a good approximation to the gradient of
the average reward, it suffices if $1/(1-\beta)$, the time constant in the
update of $z_t$, is large compared with a certain time constant---the {\it
mixing time}---of the POMDP. (It is useful, although not quite correct,
to think of the mixing time as the time from the occurrence of an action
until the effects of that action have died away.)

This gives a simple way to compute an appropriate direction to update
the parameters $\theta$. An on-line algorithm (\olpomdp) was presented
in~\cite{jair_01b} that updates the parameters $\theta$ according to
 \begin{equation}\label{equation:update}
  \theta_t = \theta_{t-1} + \gamma r_t z_t,
 \end{equation}
where the small positive real number $\gamma$ is the size of the
steps taken in parameter space.  If these steps are sufficiently small,
so that the parameters change slowly, this update rule modifies the
parameters in the direction that maximally increases the long-term
average of the reward.

\section{Direct reinforcement learning with independent agents}
\label{section:multi-drl}

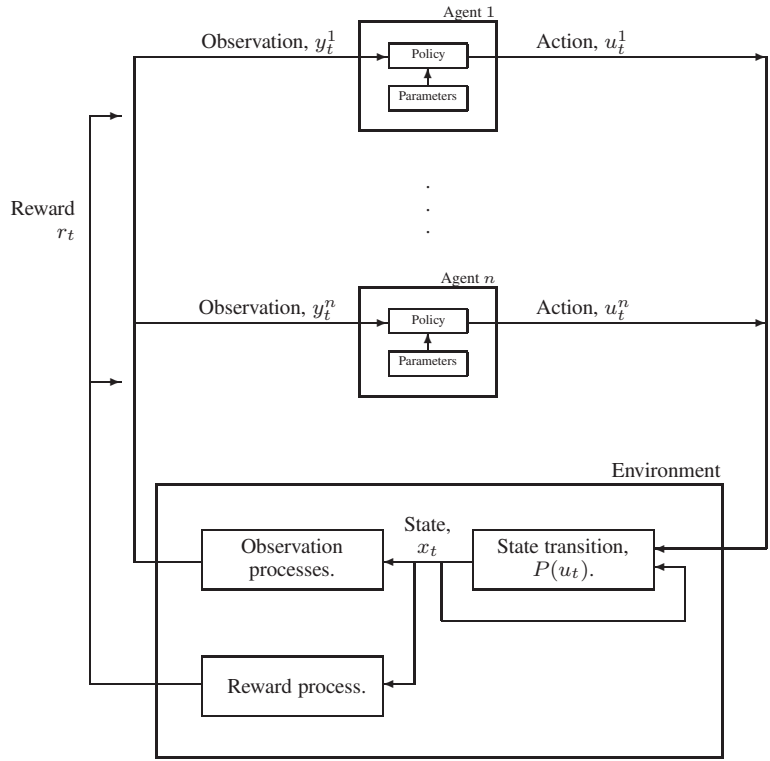
\begin{figure}
\center{\setlength{\unitlength}{2486sp}%
\begingroup\makeatletter\ifx\SetFigFont\undefined%
\gdef\SetFigFont#1#2#3#4#5{%
  \reset@font\fontsize{#1}{#2pt}%
  \fontfamily{#3}\fontseries{#4}\fontshape{#5}%
  \selectfont}%
\fi\endgroup%
\begin{picture}(7497,7533)(166,-6844)
\put(766,-1411){\makebox(0,0)[rb]{\smash{\SetFigFont{8}{9.6}{\familydefault}{\mddefault}{\updefault}Reward}}}
\put(766,-1636){\makebox(0,0)[rb]{\smash{\SetFigFont{8}{9.6}{\familydefault}{\mddefault}{\updefault}$r_t$}}}
\put(5626,-4786){\makebox(0,0)[b]{\smash{\SetFigFont{8}{9.6}{\familydefault}{\mddefault}{\updefault}State transition,}}}
\put(5626,-5011){\makebox(0,0)[b]{\smash{\SetFigFont{8}{9.6}{\familydefault}{\mddefault}{\updefault}$P(u_t)$.}}}
\thinlines
\put(4726,-5146){\framebox(1800,585){}}
\put(2926,-4786){\makebox(0,0)[b]{\smash{\SetFigFont{8}{9.6}{\familydefault}{\mddefault}{\updefault}Observation}}}
\put(2926,-5011){\makebox(0,0)[b]{\smash{\SetFigFont{8}{9.6}{\familydefault}{\mddefault}{\updefault}processes.}}}
\put(2026,-5146){\framebox(1800,585){}}
\put(2026,-6406){\framebox(1800,585){}}
\put(2971,-6181){\makebox(0,0)[b]{\smash{\SetFigFont{8}{9.6}{\familydefault}{\mddefault}{\updefault}Reward process.}}}
\put(4276,-4561){\makebox(0,0)[b]{\smash{\SetFigFont{8}{9.6}{\familydefault}{\mddefault}{\updefault}State,}}}
\put(4276,-4786){\makebox(0,0)[b]{\smash{\SetFigFont{8}{9.6}{\familydefault}{\mddefault}{\updefault}$x_t$}}}
\put(4726,-4876){\vector(-1, 0){900}}
\put(4141,-4876){\line( 0,-1){1215}}
\put(4141,-6091){\vector(-1, 0){315}}
\put(4411,-4876){\line( 0,-1){585}}
\put(4411,-5461){\line( 1, 0){2430}}
\put(6841,-5461){\line( 0, 1){540}}
\put(6841,-4921){\vector(-1, 0){315}}
\thicklines
\put(1576,-6811){\framebox(5625,2700){}}
\put(7201,-4021){\makebox(0,0)[rb]{\smash{\SetFigFont{8}{9.6}{\familydefault}{\mddefault}{\updefault}Environment}}}
\thinlines
\put(4276,-132){\vector( 0, 1){193}}
\put(3890,-363){\framebox(772,231){}}
\put(3890, 61){\framebox(772,231){}}
\put(4281,152){\makebox(0,0)[b]{\smash{\SetFigFont{5}{6.0}{\familydefault}{\mddefault}{\updefault}Policy}}}
\put(4278,-264){\makebox(0,0)[b]{\smash{\SetFigFont{5}{6.0}{\familydefault}{\mddefault}{\updefault}Parameters}}}
\put(4276,-2780){\vector( 0, 1){193}}
\put(3890,-3011){\framebox(772,231){}}
\put(3890,-2587){\framebox(772,231){}}
\put(4281,-2496){\makebox(0,0)[b]{\smash{\SetFigFont{5}{6.0}{\familydefault}{\mddefault}{\updefault}Policy}}}
\put(4278,-2912){\makebox(0,0)[b]{\smash{\SetFigFont{5}{6.0}{\familydefault}{\mddefault}{\updefault}Parameters}}}
\put(4141,-5416){\makebox(2.6458,18.5208){\SetFigFont{5}{6}{\rmdefault}{\mddefault}{\updefault}.}}
\put(7651,164){\line( 0,-1){4905}}
\put(7651,-4741){\vector(-1, 0){1125}}
\put(2026,-4876){\line(-1, 0){675}}
\put(1351,-4876){\line( 0, 1){5040}}
\put(4681,-2491){\vector( 1, 0){2970}}
\put(1351,164){\vector( 1, 0){2520}}
\put(1351,-2491){\vector( 1, 0){2520}}
\put(2026,-6091){\line(-1, 0){1125}}
\put(901,-6091){\line( 0, 1){5670}}
\put(901,-421){\vector( 1, 0){315}}
\put(901,-3076){\vector( 1, 0){315}}
\thicklines
\put(3601,-3211){\framebox(1350,1061){}}
\put(3601,-556){\framebox(1350,1061){}}
\thinlines
\put(4681,164){\vector( 1, 0){2970}}
\put(5356,-2401){\makebox(0,0)[lb]{\smash{\SetFigFont{8}{9.6}{\familydefault}{\mddefault}{\updefault}Action, $u_t^n$}}}
\put(5356,254){\makebox(0,0)[lb]{\smash{\SetFigFont{8}{9.6}{\familydefault}{\mddefault}{\updefault}Action, $u_t^1$}}}
\put(3376,254){\makebox(0,0)[rb]{\smash{\SetFigFont{8}{9.6}{\familydefault}{\mddefault}{\updefault}Observation, $y_t^1$}}}
\put(3376,-2401){\makebox(0,0)[rb]{\smash{\SetFigFont{8}{9.6}{\familydefault}{\mddefault}{\updefault}Observation, $y_t^n$}}}
\put(4276,-1141){\makebox(0,0)[b]{\smash{\SetFigFont{8}{9.6}{\familydefault}{\mddefault}{\updefault}.}}}
\put(4276,-1366){\makebox(0,0)[b]{\smash{\SetFigFont{8}{9.6}{\familydefault}{\mddefault}{\updefault}.}}}
\put(4276,-1591){\makebox(0,0)[b]{\smash{\SetFigFont{8}{9.6}{\familydefault}{\mddefault}{\updefault}.}}}
\put(4951,569){\makebox(0,0)[rb]{\smash{\SetFigFont{6}{7.2}{\familydefault}{\mddefault}{\updefault}Agent $1$}}}
\put(4951,-2086){\makebox(0,0)[rb]{\smash{\SetFigFont{6}{7.2}{\familydefault}{\mddefault}{\updefault}Agent $n$}}}
\end{picture}}
\caption{\label{figure:multi}%
POMDP controlled by $n$ independent agents.}
\end{figure}

Suppose that, instead of a single agent, there are $n$ independent
agents, all cooperating to maximize the average reward (see
Figure~\ref{figure:multi}). Suppose that each of these agents sees a
distinct observation vector, and has a distinct parameterized randomized
policy that depends on its own set of parameters. This multi-agent reinforcement
learning problem can also be modelled as a POMDP by considering this
collection of agents as a single agent, with an observation vector that
consists of the $n$ observation vectors of each independent agent, and
similarly for the parameter vector and action vector. For example, if
the $n$ agents are cooperating to clean the floor in an office, the state
vector would include the location and orientation of the $n$ agents, the
observation vector for agent $i$ might consist of the visual information
available at that agent's current location, and the actions chosen by
all $n$ agents determine the state vector at the next time instant. The
following decomposition theorem follows from a simple calculation.

\begin{theorem}\label{theorem:multi}
For a POMDP controlled by multiple independent agents, the direct
reinforcement learning update equations~(\ref{equation:zupdate})
and~(\ref{equation:update}) for the combined agent are equivalent to
those that would be used by each agent if it ignored the existence of
the other agents.

That is, if we let $y_t^i$ denote the observation vector for agent
$i$, $u_t^i$ denote the action it takes, and $\theta^i$ denote
its parameter vector, then the update equation~(\ref{equation:update})
is equivalent to the system of $n$ update equations,
 \begin{equation}\label{equation:multi-update}
  \theta_t^i = \theta_{t-1}^i + \gamma r_t z_t^i,
 \end{equation}
where the vectors $z_t^1,\ldots,z_t^n \in\Re^k$ are updated
according to
  \begin{equation}\label{equation:multi-zupdate}
   z_{t+1}^i = \beta z_t^i +
	\frac{\nabla \mu_{u_t^i}\(y_t^i,\theta^i\)}
		{\mu_{u_t^i}\(y_t^i,\theta^i\)}.
  \end{equation}
Here, $\nabla$ denotes the gradient with respect to the agent's parameters
$\theta^i$.
\end{theorem}

Effectively, each agent treats the other agents as a part of the
environment, and can update its own behaviour while remaining oblivious
to the existence of the other agents. The only communication that
occurs between these cooperating agents is via the globally distributed
reward, and via whatever influence agents' actions have on other agents'
observations.  Nonetheless, in the space of parameters of all $n$ agents,
the updates~(\ref{equation:multi-update}) adjust the complete parameter
vector (the concatenation of the vectors $\theta^i$) in the direction
that maximally increases the average reward. We shall see in the next
section that this convenient property leads to a synaptic update rule
for spiking neurons that involves only local quantities, plus a global
reward signal.

\section{Direct reinforcement learning in neural networks}
\label{section:drl-nn}

\begin{figure}
\center{\input{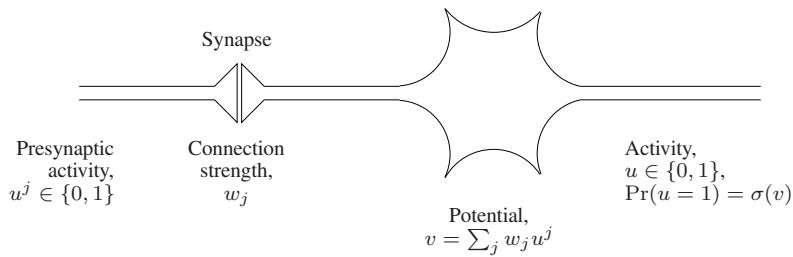}}
\caption{\label{figure:cell}%
Model of a neuron.}
\end{figure}

This section shows how we can model a neural network as a collection
of agents solving a reinforcement learning problem, and apply the
direct reinforcement learning algorithm to optimize the parameters of
the network.  The networks we consider contain simple models of spiking
neurons (see Figure~\ref{figure:cell}).  We consider discrete time, and
suppose that each neuron in the network can choose one of two actions at
time step $t$: to fire, or not to fire. We represent these actions with
the notation $u_t=1$ and $u_t=0$, respectively\footnote{The actions can be
represented by any two distinct real values, such as $u_t\in\{\pm1\}$. An
essentially identical derivation gives a similar update rule.}. We use
a simple probabilistic model for the behaviour of the neuron. Define
the {\it potential} $v_t$ in the neuron at time $t$ as
 \begin{equation}\label{equation:potential}
  v_t = \sum_j w_j u_{t-1}^j,
 \end{equation}
where $w_j$ is the connection strength of the $j$th synapse and
$u_{t-1}^j$ is the activity at the previous time step of the presynaptic
neuron at the $j$th synapse. The potential $v$ represents the voltage
at the cell body (the postsynaptic potentials having been combined
in the dendritic tree).  The probability of activity in the neuron
is a function of the potential $v$. A squashing function $\sigma$
maps from the real-valued potential to a number between $0$ and $1$,
and the activity $u_t$ obeys the following probabilistic rule.
 \begin{equation}\label{equation:neuron}
  \Pr\(\mbox{neuron fires at time $t$}\) = \Pr\(u_t=1\)
   = \sigma\(v_t\).
 \end{equation}
We assume that the squashing function satisfies
$\sigma(\alpha)=1/(1+e^{-\alpha})$.

We are interested in computation in networks of these spiking neurons,
so we need to specify the network inputs, on which the computation is
performed, and the network outputs, where the results of the computation
appear. To this end, some neurons in the network are distinguished as
{\em input neurons}, which means their activity $u_t$ is provided as an
external input to the network. Other neurons are distinguished as {\em
output neurons}, which means their activity represents the result of a
computation performed by the network.

A real-valued global reward signal $r_t$ is broadcast to every neuron in
the network at time $t$. We view each (non-input) neuron as an independent
agent in a reinforcement learning problem.  The agent's (neuron's)
policy is simply how it chooses to fire given the activities on its
presynaptic inputs. The synaptic strengths
($w_j$) are the adjustable parameters of this policy. Theorem~\ref{theorem:multi}
shows how to update the synaptic strengths in the direction that maximally
increases the long-term average of the reward. In this case, we have
 \begin{eqnarray*}
  \frac{ \frac{\partial}{\partial w_j} \mu_{u_t}}{\mu_{u_t}}
   & = & \left\{ \begin{array}{ll}
	\frac{\sigma'\(v_t\) u_{t-1}^j}
		{\sigma\(v_t\)} &
			\mbox{if $u_t=1$,} \\[4mm]
	\frac{-\sigma'\(v_t\) u_{t-1}^j}
		{1-\sigma\(v_t\)} &
			\mbox{otherwise}
	\end{array}\right. \\
   & = & \( u_t - \sigma\(v_t\) \) u_{t-1}^j,
 \end{eqnarray*}
where the second equality follows from the property of the
squashing function,
 $$
  \sigma'(\alpha) = \sigma(\alpha)\(1-\sigma(\alpha)\).
 $$
This results in an update rule for the $j$-th synaptic strength of
 \begin{equation}\label{equation:neuron-update}
  w_{j,t+1} = w_{j,t} + \gamma r_{t+1} z_{j,t+1},
 \end{equation}
where the real numbers $z_{j,t}$ are updated according to
  \begin{equation}\label{equation:neuron-zupdate}
   z_{j,t+1} = \beta z_{j,t} +
	\( u_t - \sigma\(v_t\) \) u_{t-1}^j.
  \end{equation}
These equations describe the updates for the parameters in a single
neuron.  The pseudocode in Algorithm~\ref{algorithm:net} gives a complete
description of the steps involved in computing neuron activities and
synaptic modifications for a network of such neurons.

\begin{algorithm}
\caption{Model of neural network activity and synaptic modification.}
\label{algorithm:net}
\begin{algorithmic}[1]
\STATE {\bf Given: } \\[2mm]
  \begin{tabular}{l}
  Coefficient $\beta \in [0,1)$,\\
  Step size $\gamma$, \\
  Initial synaptic connection strengths of the $i$-th neuron $w_{j,0}^i$.
  \end{tabular}
\FOR{time $t=0,1,\dots$}
	\STATE Set activities $u_t^j$ of input neurons.
	\FOR{non-input neurons $i$}
		\STATE Calculate potential $v_{t+1}^i = \sum_j w_{j,t}^i u_t^j$.
		\STATE Generate activity $u_{t+1}^i\in\{0,1\}$ using
			$\Pr\(u_{t+1}^i=1\) = \sigma\(v_{t+1}^i\)$.
	\ENDFOR
	\STATE Observe reward $r_{t+1}$ (which depends on network outputs).
	\FOR{non-input neurons $i$}
		\STATE Set $z_{j,t+1}^i = \beta z_{j,t+1}^i +
			\( u_t^i - \sigma\(v_t^i\) \) u_{t-1}^j$.\\[2mm]
		\STATE Set $w_{j,t+1}^i = w_{j,t}^i +
			\gamma r_{t+1} z_{j,t+1}^i$.
	\ENDFOR
\ENDFOR
\end{algorithmic}
\end{algorithm}

Suitable values for the quantities $\beta$ and $\gamma$ required by
Algorithm~\ref{algorithm:net} depend on the mixing time of the controlled
POMDP.  The coefficient $\beta$ sets the decay rate of the variable
$z_t$. For the algorithm to accurately approximate the gradient direction,
the corresponding time constant, $1/(1-\beta)$, should be large compared
with the mixing time of the environment. The step size $\gamma$ affects
the rate of change of the parameters. When the parameters are constant,
the long term average of $r_tz_t$ approximates the gradient. Thus, the
step size $\gamma$ should be sufficiently small so that the parameters
are approximately constant over a time scale that allows an accurate
estimate. Again, this depends on the mixing time. Loosely speaking,
both $1/(1-\beta)$ and $1/\gamma$ should be significantly larger than
the mixing time.

\section{Biological Considerations}
\label{section:mechanisms}

\begin{figure}
\center{\input{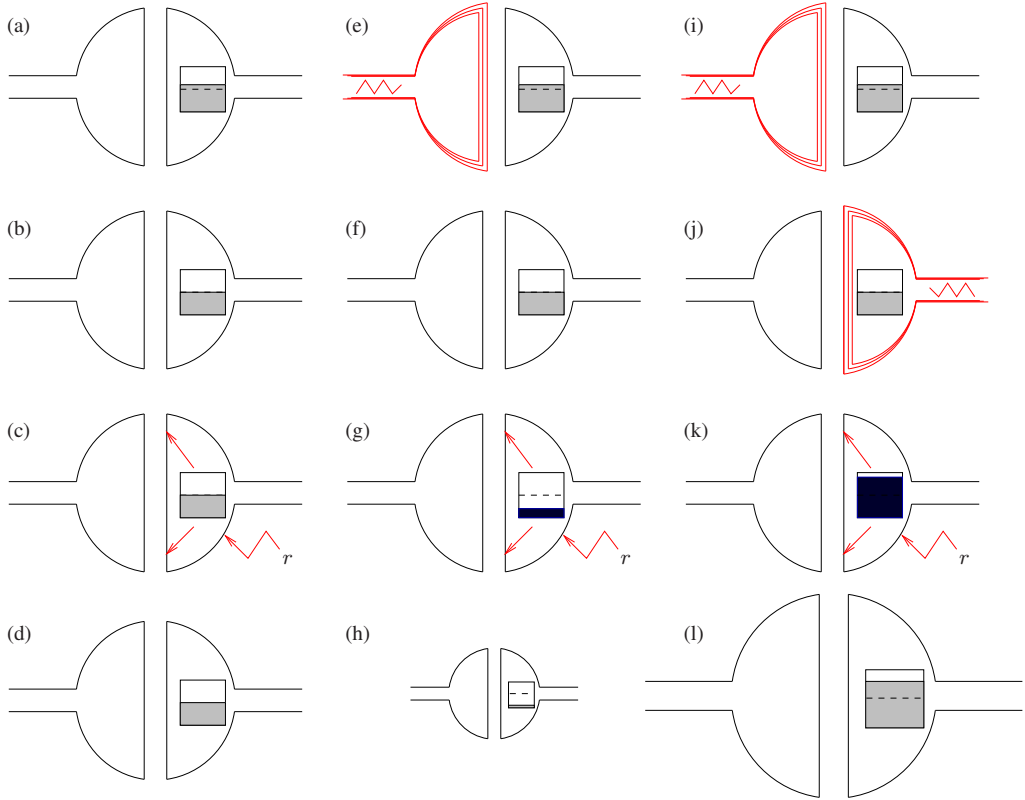}}
 \caption{\label{figure:synapse}%
   An illustration of synaptic updates. The presynaptic neuron is
   on the left, the postsynaptic on the right. The level inside
   the square in the postsynaptic neuron represents the quantity
   $z_j$. (The dashed line indicates the zero value.) The symbol $r$
   represents the presence of a positive value of the reward signal,
   which is assumed to take only two values here.  (The postsynaptic
   location of $z_j$ and $r$ is for convenience in the depiction, and
   has no other significance.)  The size of the synapse represents the
   connection strength. Time proceeds from top to bottom. (a)--(d):
   A sequence through time illustrating changes in the synapse when
   no action potentials occur.  In this case, $z_j$ steadily decays
   ((a)--(b)) towards zero, and when a reward signal arrives (c), the
   strength of the synaptic connection is not significantly adjusted.
   (e)--(h): Presynaptic action potential (e), but no postsynaptic action
   potential (f) leads to a larger decrease in $z_j$ (g), and subsequent
   decrease in connection strength on arrival of the reward signal (h).
   (i)--(l): Presynaptic action potential (i), followed by postsynaptic
   action potential (j) leads to an increase in $z_j$ (k) and subsequent
   increase in connection strength (l).}
\end{figure}

In modifying the strength of a synaptic connection, the update
rule described by Equations~(\ref{equation:neuron-update})
and~(\ref{equation:neuron-zupdate}) involves two components
(see Figure~\ref{figure:synapse}). There is a Hebbian component
($u_tu_{t-1}^j$) that helps to increase the synaptic connection strength
when firing of the postsynaptic neuron follows firing of the presynaptic
neuron.  When the firing of the presynaptic neuron is not followed by
postsynaptic firing, this component is $0$, while the second component
($- \sigma\(v_t\) u_{t-1}^j$) helps to decrease the synaptic connection
strength.

The update rule has several attractive properties.
 \begin{description}
  \item[Locality] The modification of a particular synapse $w_j$ involves
  the postsynaptic potential $v$, the postsynaptic activity $u$, and
  the presynaptic activity $u^j$ at the previous time step.

  Certainly the postsynaptic potential is available at the synapse.
  Action potentials in neurons are transmitted back up the dendritic
  tree~\cite{stuart94}, so that (after some delay) the postsynaptic
  activity is also available at the synapse. Since the influence of
  presynaptic activity on the postsynaptic potential is mediated by
  receptors at the synapse, evidence of presynaptic activity is also
  available at the synapse. While Equation~(\ref{equation:neuron-zupdate})
  requires information about the {\it history} of presynaptic activity,
  there is some evidence for mechanisms that allow recent receptor
  activation to be remembered~\cite{Markram97,BiPoo98}.  Hence, all of
  the quantities required for the computation of the variable $z_j$
  are likely to be available in the postsynaptic region.

  \item[Simplicity] The computation of $z_j$
  in~(\ref{equation:neuron-zupdate}) involves only additions and
  subtractions modulated by the presynaptic and postsynaptic activities,
  and combined in a simple first order filter. This filter is a leaky
  integrator which models, for instance, such common features as the
  concentration of ions in some region of a cell or the potential across
  a membrane. Similarly, the connection strength updates described by
  Equation~(\ref{equation:neuron-update}) involve simply the addition
  of a term that is modulated by the reward signal.

  \item[Optimality] The results from~\cite{jair_01a}, together with
  Theorem~\ref{theorem:multi}, show that this simple update rule modifies
  the network parameters in the direction that maximally increases the
  average reward, so it leads to parameter values that locally optimize
  the performance of the network.
 \end{description}

There are some experimental results that are consistent
with the involvement of the correlation component (the term
$(u_t-\sigma(v_t))u_{t-1}^j$) in the parameter updates. For instance,
a large body of literature on long-term potentiation (beginning
with~\cite{BlissLomo73}) describes the enhancement of synaptic efficacy
following association of presynaptic and postsynaptic activities. More
recently, the importance of the relative timing of the EPSPs and
APs has been demonstrated~\cite{Markram97,BiPoo98}. In particular,
the postsynaptic firing must occur after the EPSP for enhancement to
occur. The backpropagation of the action potential up the dendritic tree
appears to be crucial for this~\cite{MageeJohnston97}.

There is also experimental evidence that presynaptic activity without
the generation of an action potential in the postsynaptic cell can
lead to a decrease in the connection
strength~\cite{StantonSejnowski89}. The recent
finding~\cite{Markram97,BiPoo98} that
an EPSP occurring shortly {\em after} an AP can lead to depression is
also consistent with this aspect of Hebbian learning. However, in the
experiments reported in~\cite{Markram97,BiPoo98}, the presence of the
AP appeared to be important. It is not clear if the significance of
the relative timings of the EPSPs and APs is related to learning or to
maintaining stability in bidirectionally coupled cells.

Finally, some experiments have demonstrated a decrease in synaptic
efficacy when the synapses were not involved in the production of an
action potential~\cite{LoPoo91}.

The update rule also requires a reward signal that is broadcast to all
neurons in the network.  In all of the experiments mentioned above,
the synaptic modifications were observed without any evidence of
the presence of a plausible reward signal.  However, there is limited
evidence for such a signal in brains. It could be delivered in the form
of particular neurotransmitters, such as serotonin or nor-adrenaline, to
all neurons in a circuit.  Both of these neurotransmitters are delivered
to the cortex by small cell assemblies (the raphe nucleus and the locus
coeruleus, respectively) that innervate large regions of the cortex. The
fact that these assemblies contain few cell bodies suggests that they
carry only limited information. It may be that the reward signal is
transmitted first electrically from one of these cell assemblies,
and then by diffusion of the neurotransmitter to all of the plastic
synaptic connections in a neural circuit. This would save the expense
of a synapse delivering the reward signal to every plastic connection,
but could be significantly slower.  This need not be a disadvantage; for
the purposes of parameter optimization, the required rate of delivery of
the reward signal depends on the time constants of the task, and can be
substantially slower than cell signalling times.  There is evidence that
the local application of serotonin immediately after limited synaptic
activity can lead to long term facilitation~\cite{ClarkKandel93}.

\section{Simulation Results}
\label{section:simulation}

In this section, we describe the results of simulations of
Algorithm~\ref{algorithm:net} for a pattern classification problem
and an adaptive control problem. In all simulation experiments, we
used a symmetric representation, $u\in\{-1,1\}$. The difference
between this representation and the assymmetric $u\in\{0,1\}$ is
a simple transformation of the parameters, but this can be significant
for gradient descent procedures.

\subsection{Sonar signal classification}

Algorithm~\ref{algorithm:net} was applied to the problem of sonar return
classification studied by Gorman and Sejnowski~\cite{GorSej88}.  (The data
set is available from the U.~C.~Irvine repository~\cite{UCI}.)  Each
pattern consists of 60 real numbers in the range $[0,1]$, representing
the energy in various frequency bands of a sonar signal reflected from
one of two types of underwater objects, rocks and metal cylinders. The
data set contains 208 patterns, 97 labeled as rocks and 111 as cylinders.
We investigated the performance of a two-layer network of spiking neurons
on this task. The first layer of $8$ neurons received the vector of 60
real numbers as inputs, and a single output neuron received the outputs
of these neurons. This neuron's output at each time step was viewed
as the prediction of the label corresponding to the pattern presented
at that time step. The reward signal was $0$ or $1$, for an incorrect
or correct prediction, respectively. The parameters of the algorithm
were $\beta=0.5$ and $\gamma=10^{-4}$. Weights were initially set to
random values uniformly chosen in the interval $(-0.1,0.1)$. Since it
takes two time steps for the influence of the hidden unit parameters
to affect the reward signal, it is essential for the value of $\beta$
for the synapses in a hidden layer neuron to be positive.  It can be
shown that for a constant pattern vector, the optimal choice of $\beta$
for these synapses is $0.5$.

Each time the input pattern changed, the delay through the network
meant that the prediction corresponding to the new pattern was delayed by
one time step. Because of this, in the experiments each pattern was
presented for many time steps before it was changed.

\begin{figure}
\center{
\includegraphics[scale=0.8]{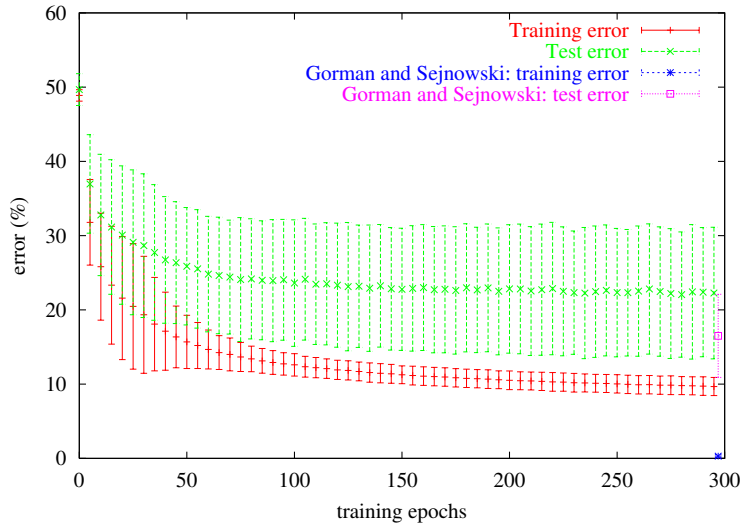}
}
 \caption{\label{figure:sonar}%
   Learning curves for the sonar classification problem.}
\end{figure}

Figure~\ref{figure:sonar} shows the mean and standard deviation of
training and test errors over 100 runs of the algorithm plotted against
the number of training epochs. Each run involved an independent random
split of the data into a test set (10\%) and a training set (90\%).
For each training epoch, patterns in the training set were presented to
the network for $1000$ time steps each. The errors were calculated as the
proportion of misclassifications during one pass through the data, with
each pattern presented for $1000$ time steps.  Clearly, the algorithm
reliably leads to parameter settings that give training error around
$10\%$, without passing any gradient information through the network.

Gorman and Sejnowski~\cite{GorSej88} investigated the performance of
sigmoidal neural networks on this data. Although the networks they
used were quite different (since they involved deterministic units
with real-valued outputs), the training error and test error they
reported for a network with 6 hidden units is also illustrated in
Figure~\ref{figure:sonar}.

\subsection{Controlling an inverted pendulum}

\begin{figure}
\center{
\includegraphics[scale=1]{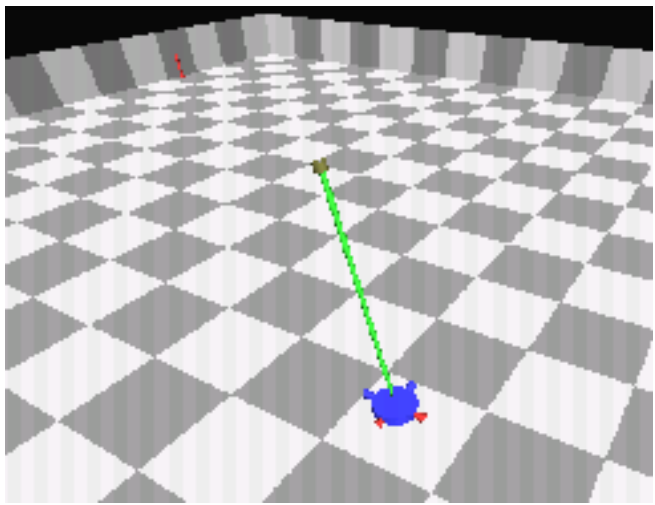}
}
 \caption{\label{figure:pendulum}%
   The inverted pendulum.}
\end{figure}

We also considered a problem of learning to balance an inverted
pendulum. Figure~\ref{figure:pendulum} shows the arrangement: a puck
moves in a square region. On the top of the puck is a weightless rod
with a weight at its tip. The puck has no internal dynamics.

We investigated the performance of Algorithm~\ref{algorithm:net} on this
problem. We used a network with four hidden units, each receiving real
numbers representing the position and velocity of the puck and the angle
and angular velocity of the pendulum. These units were connected to two
more units, whose outputs were used to control the sign of two 10N thrusts
applied to the puck in the two axis directions. The reward signal was $0$
when the pendulum was upright, and $-1$ when it hit the ground. Once the
pendulum hit the ground, the puck was randomly located near the centre of
the square with velocity zero, and the pendulum was reset to vertical
with zero angular velocity.

In the simulation, the square was $5\times 5$ metres, the dynamics
were simulated in discrete time, with time steps of $0.02$s, the puck
bounced elastically off the walls, gravity was $9.8$ms$^{-2}$, the
puck radius was $50$mm, the puck height was $0$, the puck mass was
$1$kg, air resistance was neglected, the pendulum length was $500$mm,
the pendulum mass was $100$g, the coefficient of friction of the puck
on the ground was $5\times 10^{-4}$, and friction at the pendulum
joint was set to zero.

The algorithm parameters were $\gamma=10^{-6}$ and $\beta=0.995$.

\begin{figure}
\center{
\includegraphics[scale=0.8]{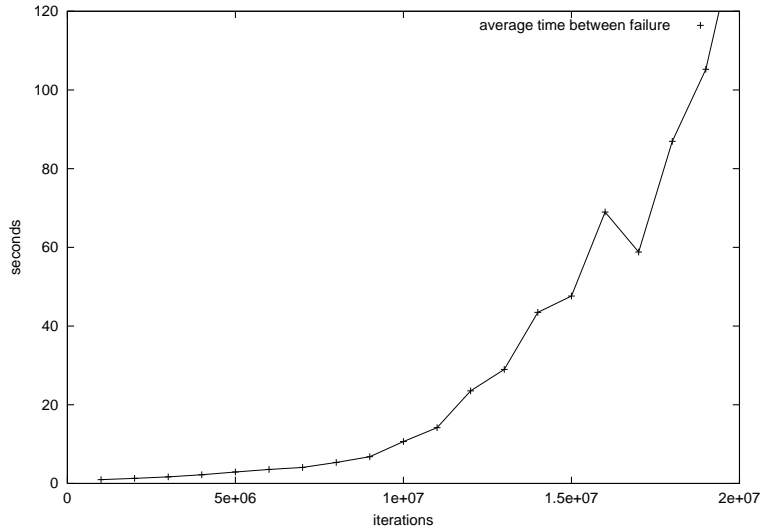}
}
 \caption{\label{figure:pendlearn}%
   A typical learning curve for the inverted pendulum problem.}
\end{figure}

Figure~\ref{figure:pendlearn} shows a typical learning curve: the
average time before the pendulum falls (in a simulation of $100 000$
iterations $= 2000$ seconds), as a function of total simulated time.
Initial weights were chosen uniformly from the interval $(-0.05,0.05)$.

\section{Further work}
\label{section:further}

The most interesting questions raised by these results are concerned
with possible biological mechanisms for update rules of this type.
Some aspects of the update rule are supported by experimental results.
Others, such as the reward signal, have not been investigated
experimentally. One obvious direction for this work is the development
of update rules for more realistic models of neurons. First, the model
assumes discrete time.  Second, it ignores some features that biological
neurons are known to possess. For instance, the location of synapses in the
dendritic tree allow timing relationships between action potentials in
different presynaptic cells to affect the resulting postsynaptic
potential.
Other features of dendritic processing, such as nonlinearities, are
also ignored by the model presented here.  It is not clear which of
these features are important for the computational properties of neural
circuits.


\section{Conclusions}
\label{section:conclusions}

The synaptic update rule presented in this paper requires only simple
computations involving only local quantities plus a global reward
signal. Furthermore, it adjusts the synaptic connection strengths
to locally optimize the average reward received by the network.
The reinforcement learning paradigm encompasses a considerable variety
of learning problems. Simulations have shown the effectiveness of the
algorithm for a simple pattern classification problem and an adaptive
control problem.

\bibliographystyle{abbrv}
\bibliography{bib}

\end{document}